\documentclass[10pt,twocolumn,letterpaper]{article}

\usepackage{cvpr}  
\usepackage{makecell}
\usepackage{multirow}
\usepackage{rotating}
\usepackage[accsupp]{axessibility}
\usepackage{tabularray}
\usepackage{graphicx}
\usepackage{multirow}
\usepackage{booktabs}
\usepackage{adjustbox}   

\newcommand{\rotcelll}[1]{\adjustbox{angle=90,lap=0.2\width}{\strut{#1}}}
\definecolor{cvprblue}{rgb}{0.21,0.49,0.74}
\usepackage[pagebackref,breaklinks,colorlinks,allcolors=cvprblue]{hyperref}

\def\paperID{19046} 
\def\confName{CVPR}
\def\confYear{2026}

\title{Modality-Aware and Anatomical Vector-Quantized Autoencoding for Multimodal Brain MRI}
\author{
    Mingjie Li \quad Edward Kim \quad Yue Zhao \quad Ehsan Adeli \quad Kilian M. Pohl\thanks{Corresponding author. Emails are \textit{\{lmj695, kpohl\}@stanford.edu} \\ Code is available at \url{https://github.com/mlii0117/NeuroQuant}} \\
     Stanford University
}

\begin{document}
\maketitle
\begin{abstract}
Learning a robust Variational Autoencoder (VAE) is a fundamental step for many deep learning applications in medical image analysis, such as MRI synthesizes. Existing brain VAEs predominantly focus on single-modality data (i.e., T1-weighted MRI), overlooking the complementary diagnostic value of other modalities like T2-weighted MRIs. Here, we propose a modality-aware and anatomically grounded 3D vector-quantized VAE (VQ-VAE) for reconstructing multi-modal brain MRIs. Called NeuroQuant, it first learns a shared latent representation across modalities using factorized multi-axis attention, which can capture relationships between distant brain regions. It then employs a dual-stream 3D encoder that explicitly separates the encoding of modality-invariant anatomical structures from modality-dependent appearance. Next, the anatomical encoding is discretized using a shared codebook and combined with modality-specific appearance features via Feature-wise Linear Modulation (FiLM) during the decoding phase. This entire approach is trained using a joint 2D/3D strategy in order to account for the slice-based acquisition of 3D MRI data. Extensive experiments on two multi-modal brain MRI datasets demonstrate that NeuroQuant achieves superior reconstruction fidelity compared to existing VAEs, enabling a scalable foundation for downstream generative modeling and cross-modal brain image analysis.
\end{abstract}

\section{Introduction}
\label{sec:intro}

Multimodal structural MRI is central to the diagnosis of many brain diseases.
Complementary contrasts (e.g., T1, T2, and FLAIR) enable a comprehensive assessment of brain anatomy, tissue integrity, and pathology~\cite{miller2016multimodal,wen2024genetic}.
However, acquiring high-quality multimodal MRI for every subject remains challenging due to motion artifacts, time constraints, and cost limitations, often resulting in limited or imbalanced datasets.
To address these limitations, brain MRI synthesis~\cite{wang2025toward,wang2025self,peng2024latent,peng2023cDPM,peng2024brainsyn,monaildm,singh2021medical,puglisi2025brain} has emerged as a promising direction for generating high-quality and anatomically plausible 3D scans across modalities.
Such synthesized MRIs not only facilitate data augmentation and improve the robustness of downstream tasks~\cite{dayarathna2025mu}, but also enable counterfactual generation to support causal analysis and disease mechanism exploration~\cite{cf_mri,yeganeh2025latent,Pug_Enhancing_MICCAI2024,li2025integrating}.

Some popular approaches for synthesis are based on latent diffusion models (LDM)~\cite{rombach2022high}.
They first employ a Variational Autoencoder (VAE)~\cite{kingma2013auto} to compress high-dimensional 3D MRI volumes into a compact latent space, where the subsequent diffusion process~\cite{ho2020denoising} operates more efficiently.
The VAEs used in these pipelines can generally be grouped into two types. The first trains a dataset-specific VAE from scratch (e.g., cDPM~\cite{peng2024brainsyn} and BrLP~\cite{puglisi2025brain,Pug_Enhancing_MICCAI2024}), which aligns closely with dataset idiosyncrasies but incurs heavy retraining cost and often limits cross-study generalization. The second adopts off-the-shelf 2D VAEs pretrained on natural images~\cite{rombach2022high} and applies them slice-by-slice to brain volumes~\cite{yeganeh2025latent}. Because these models operate independently on individual slices and lack any mechanism to capture volumetric context or distinguish MRI contrasts (e.g., T2, FLAIR), they frequently produce inconsistent anatomical boundaries across orientations and modality-biased reconstructions.

\begin{figure}[t!]
    \centering
    \includegraphics[width=1\linewidth]{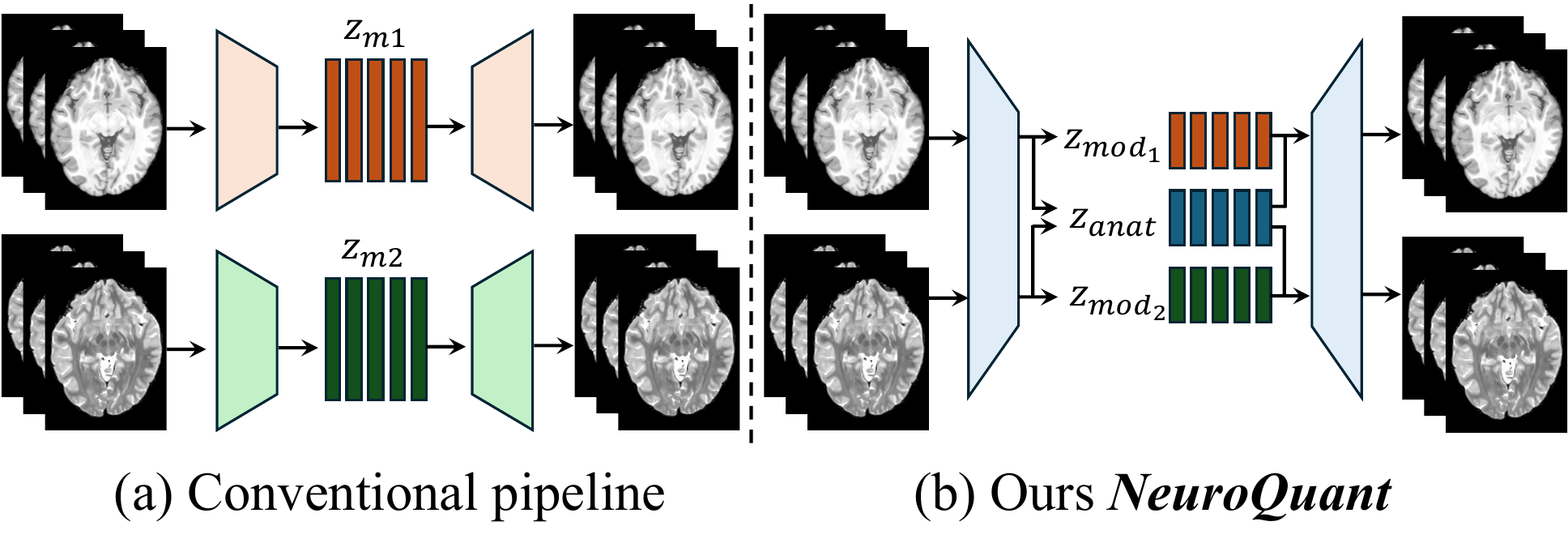}
    \caption{Conventional approaches handle different MRI modalities by duplicating separate encoders. In contrast, we employ a unified 3D encoder that disentangles shared anatomical representations $z_{anat}$ from modality-specific features $z_{mod}$ within the same latent space.}
    \label{fig:motivation}
\end{figure}

A remedy is to learn a single unified brain VAE that can jointly model anatomical structure and modality-specific appearance while preserving volumetric coherence and scalability. Recent progress in large, general-purpose VAEs has shown that training a shared encoder–decoder on heterogeneous medical images (e.g., CT, MRI, X-ray) can yield broadly reusable latent representations~\cite{ma2025meditok,varma2025medvae}.
However, these ``universal medical VAEs'', such as MedVAE~\cite{varma2025medvae} and MediTok~\cite{ma2025meditok}, derive their generality primarily from cross-organ and cross-modality data diversity, rather than from explicitly modeling the complementary contrasts within multi-modal brain MRI. In practice, their training strategies treat T1-weighted, T2-weighted, and other MRI sequences either as independent inputs requiring separate encoders, or are indifferent to differences in modality  by passing each MRI through a single backbone. As a result, while these models are effective as general medical tokenizers, they do not exploit the rich correspondence among brain MRI modalities with respect to brain anatomy. To overcome this issue, we propose here a multimodal 3D brain VAE that explicitly disentangles shared anatomical structure from modality-specific appearance (\textit{e.g.}, local intensity distributions and contrast characteristics), enabling coherent cross-modal representation learning and faithful 3D MRI reconstruction. 

Specifically, NeuroQuant is unified multimodal 3D VQ-VAE designed for anatomically faithful and modality-consistent brain MRI representation. NeuroQuant operates directly in 3D volumetric space, enabling coherent modeling of brain structures across modalities within a shared latent representation. As illustrated in \Cref{fig:motivation}, unlike conventional pipelines that duplicate modality-specific encoders, NeuroQuant adopts a dual-stream architecture that explicitly disentangles modality-invariant anatomical information from modality-dependent appearance cues. The first stream encodes spatially consistent brain structure and is discretized through a shared vector-quantized codebook, forming a compact vocabulary of anatomical tokens. Meanwhile, the second stream captures intensity and contrast variations unique to each MRI modality and dynamically modulates the decoder via Feature-wise Linear Modulation (FiLM)~\cite{perez2018film}. To further enhance structural coherence and long-range dependency modeling, factorized multi-axis attention blocks are incorporated throughout the encoder and decoder, enabling context-aware information exchange across slices and modalities. Finally, we adopt a 2D/3D joint training strategy, allowing NeuroQuant to leverage both full volumetric scans and individual slices, improving scalability without compromising anatomical fidelity.

We evaluate NeuroQuant on two large-scale multimodal brain MRI datasets (NCANDA~\cite{ncanda} and ABCD~\cite{abcd}) using three complementary evaluation protocols: voxel-level reconstruction quality (PSNR, SSIM), structure-level anatomical consistency, and the utility of the latent representation utility for sex-classification. NeuroQuant consistently outperforms existing VAEs, demonstrating its ability to generate high-fidelity and anatomically plausible brain MRIs while producing compact and discriminative latent representations suitable for downstream analysis. In summary, our main contributions are threefold:
\begin{itemize}
    \item We propose NeuroQuant, a 3D VQ autoencoding framework for multimodal brain MRI that jointly captures anatomical structure and modality-specific appearance in a shared latent space. 
    \item We introduce a dual-stream encoder with VQ anatomical tokens and a modality-FiLM decoder, reinforced by the factorized multi-axis attention for consistent structural encoding across slices and modalities.
    \item Extensive experiments on NCANDA and ABCD demonstrate that NeuroQuant achieves state-of-the-art reconstruction fidelity, stronger structural consistency, and more discriminative latent representations.
\end{itemize}

\section{Related Work}\label{sec:related work}

\subsection{VAEs in Brain Latent Diffusion Models}

Recent LDM-based brain synthesis methods~\cite{peng2024brainsyn,puglisi2025brain,monaildm} employ a VAE to encode MRI volumes into a low-dimensional latent space.
These VAEs are typically trained separately for each dataset to align with its specific intensity and anatomical characteristics. For example, Pinaya \textit{et al.}~\cite{monaildm} trained a 3D convolutional VAE optimized with a combination of an L1 loss, perceptual loss, a patch-based adversarial objective, and a KL regularization on the latent space to improve reconstruction fidelity and stability. A similar setting was later adopted by \cite{puglisi2025brain,Pug_Enhancing_MICCAI2024}. In parallel, Peng \textit{et al.}~\cite{peng2024brainsyn} proposed a 3D VQ-VAE tailored for T1-weighted MRIs. To address the coarse quantization artifacts of the standard VQ implementation~\cite{esser2021taming}, they introduced a fine-grained hierarchical quantization strategy: each latent vector is first mapped to its nearest codebook entry and then refined by an additional quantization of the residual, effectively stacking two levels of quantized representations to improve anatomical detail preservation in 3D volumes. This VQ-VAE is also applied in the following work~\cite{peng2024latent} for counterfactual brain synthesis. Beyond directly encoding the entire 3D volume, Li \textit{et al.}~\cite{li2025integrating} utilized a causal encoder~\cite{chen2024od}, which differs from prior LDM-based approaches by operating at the 2D slice level. Inspired by video VAEs~\cite{xing2024large}, this causal encoder models inter-slice dependencies by treating each MRI slice as a sequential frame while maintaining cross-slice consistency. Although these VAEs produce compact latent representations for efficient diffusion, they remain dataset-specific and require retraining for each new dataset, thereby increasing computational cost and limiting generalization across studies. More recently, Yousef \textit{et al.}~\cite{yeganeh2025latent} employed the pretrained VAE from Stable Diffusion (SD)~\cite{rombach2022high} originally trained on large-scale natural image datasets. This pretrained VAE enables efficient latent compression and facilitates counterfactual brain MRI generation with promising quality. However, due to the substantial domain gap between natural and medical images, particularly in structural regularity and intensity distribution, its representations remain suboptimal for brain MRI, underscoring the need for a unified VAE specifically designed for neuroimaging data.

\subsection{Unified VAEs in Natural and Medical Imaging}

The idea of developing unified and reusable VAEs has gained increasing attention in both natural and medical imaging communities.  
In the natural image domain, large-scale VAEs~\cite{zhao2024image,xing2024large,chen2024od,zheng2024opensora,yu2024image,zhou2022magicvideo} have demonstrated strong generalization and adaptability for downstream generative tasks such as image synthesis, editing, and video generation.  
These models learn a compact, semantically meaningful latent space that can be reused across datasets and applications, enabling diffusion and autoregressive models to operate efficiently while maintaining high perceptual quality.  
More recently, the focus in the natural image community has shifted toward VQ-VAE-based models, often referred to as image tokenizers, which discretize visual content into compact codebook representations compatible with large language models (LLMs) for autoregressive generation. Such unified tokenizers not only enable high-quality image and video synthesis through sequence modeling but also facilitate multimodal understanding tasks by providing a shared multi-modal discrete latent space~\cite{ma2025towards,wang2024emu3}. While these tokenizers excel at reconstruction, they often lack explicit control over independent factors of variation. Parallel to these advances, VAEs frequently rely on disentangled representation learning (such as beta-VAE and FactorVAE~\cite{higgins2017betaVAE,kim2018factorVAE}), which encourage the latent space to separate underlying generative factors of variation.
To move beyond the rigid constraints of early VAEs, recent formulations incorporate more adaptive disentanglement mechanisms, such as domain intersection/difference decomposition~\cite{benaim2019domain} for unpaired data and causal-effect-aware models~\cite{you2025disentangled} that explicitly model the hierarchical relationships between latent variables. However, these methods are typically designed for 2D natural images or low-dimensional domains.

Several recent studies have introduced general-purpose medical VAEs trained on large-scale, multimodal medical image datasets that also include T1-weighted brain scans. MedVAE~\cite{varma2025medvae} adopts a 3D convolutional architecture to learn volumetric anatomical representations, emphasizing spatial consistency across diverse modalities. In contrast, MediTok~\cite{ma2025meditok} employs a 2D VQ-VAE framework to discretize medical images into tokenized representations for foundation modeling, enabling compatibility with autoregressive generative models. While both models demonstrate promising generalization across medical domains, their strategies for processing 3D brain data remain limited. MedVAE focuses on single-modality reconstruction, whereas MediTok lacks volumetric continuity due to its slice-wise formulation. Building upon these advances, our work aims to extend general-purpose medical VAEs toward a unified multimodal 3D brain VAE that jointly models anatomical and modality-specific representations with structural coherence.
\section{Methodology}\label{sec:method}

\begin{figure*}[!tb]
    \centering
    \includegraphics[width=0.95\linewidth]{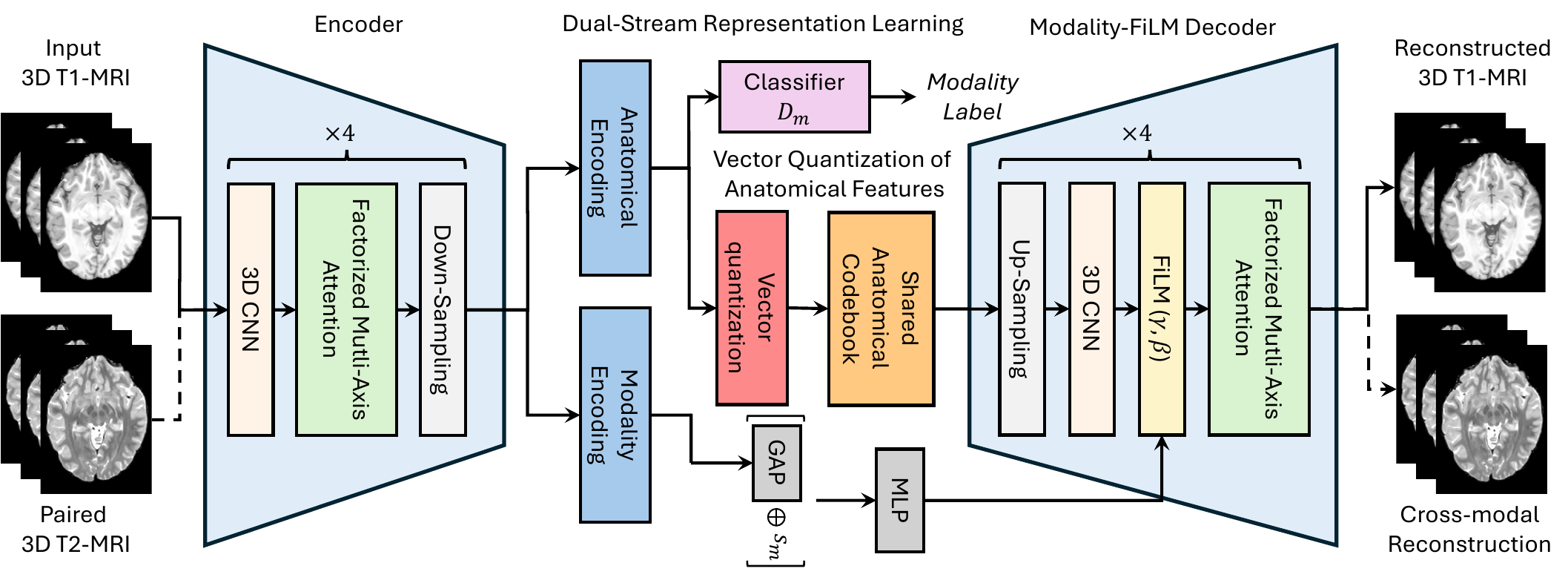}
    \caption{Architecture of NeuroQuant: It learns to disentangle anatomical and modality-specific representations from multi-modal brain MRI. The encoder uses factorized multi-axis attention to extract shared anatomical features and modality-adaptive cues. Anatomical features are discretized via a shared vector-quantized codebook, while modality features generate FiLM parameters $(\gamma, \beta)$ through a global modality embedding $s_m$ and MLP to guide the decoder. The Modality-FiLM decoder reconstructs 3D MR volumes with modality-adaptive contrast and anatomical coherence. Cross-modal reconstruction and modality prediction ensure disentanglement and consistency across modalities. $\bigoplus$ refers to the concatenate operation.}
    \vspace{-0.5cm}
    \label{fig:overview}
\end{figure*}

In this section, we present the design of NeuroQuant, illustrated in \Cref{fig:overview}, a unified dual-stream 3D VQ-VAE for multimodal brain MRI autoencoding.  The model is built upon a 3D encoder–decoder architecture augmented with factorized multi-axis attention, a shared quantized anatomical codebook, and a FiLM-based modality conditioning module to disentangle structure and appearance. In addition, a 2D/3D joint training strategy allows the model to learn coherently from both volumetric data and planar slices without introducing extra encoders. We first describe the network design, followed by the codebook quantization and joint training scheme.

\subsection{Overview}

Let $x_m \in \mathbb{R}^{D\times H\times W}$ denote a 3D brain MRI of modality $m \in {\mathrm{T1}, \mathrm{T2}}$, where $D$, $H$, and $W$ correspond to the number of slices along the axial, coronal, and sagittal dimensions, respectively. Our objective is to encode $x_m$ into disentangled latent representations that capture (1) anatomical structure shared across modalities and (2) modality-specific contrast defining appearance variations. Formally, the encoder $E(\cdot)$ consists of two parallel branches and maps the input volume into disentangled latent codes:
\begin{equation}
E(x_m) = \big(E_{\mathrm{anat}}(x_m),\,E_{\mathrm{mod}}(x_m)\big)
        = (z_{\mathrm{anat}}, z_{\mathrm{mod}}),
\end{equation}
where $z_{\mathrm{anat}}\in\mathbb{R}^{C_a\times D_a\times H_a\times W_a}$ encodes structural information, 
and $z_{\mathrm{mod}}\in\mathbb{R}^{C_m\times D_a\times H_a\times W_a}$ captures modality-dependent features.
The anatomical latent $z_{\mathrm{anat}}$ is discretized through vector quantization, 
while the modality latent $z_{\mathrm{mod}}$ modulates the decoder via FiLM~\cite{perez2018film}.

\subsection{Factorized Multi-Axis Attention}

CNNs are effective at modeling local spatial patterns but struggle to capture long-range anatomical dependencies across distant brain regions.  
A straightforward extension to incorporate global context is to apply full 3D self-attention, whose computational cost scales quadratically with the number of voxels, i.e., $\mathcal{O}((DHW)^2)$. Such complexity is prohibitive for high-resolution brain volumes.

To balance efficiency and global modeling, we design a factorized multi-axis attention encoder that decomposes 3D attention into a sequence of axis-wise operations along the axial, coronal, and sagittal dimensions. Given an intermediate feature map $F^{(t)} \in \mathbb{R}^{C^{(t)} \times D^{(t)} \times H^{(t)} \times W^{(t)}}$ at stage~$t$, the attention along axis $a \in \{\text{axial},\text{coronal},\text{sagittal}\}$ is computed as:
\begin{equation}
\mathrm{Attn}_a(F^{(t)}) =
\mathrm{Softmax}\!\left(
\frac{Q_a K_a^\top}{\sqrt{d_k}}
\right)V_a,
\end{equation}
where $Q_a, K_a, V_a \in \mathbb{R}^{(L_a \times d_k)}$ are query, key, and value projections obtained by flattening $F^{(t)}$ along axis~$a$, and $L_a$ denotes the sequence length along that axis. By performing attention sequentially across the three spatial axes, the overall complexity reduces from $\mathcal{O}((DHW)^2)$ to $\mathcal{O}(DHW(D+H+W))$, providing a substantial efficiency gain while preserving global spatial coherence.

The encoder comprises four stacked attention blocks. Each block contains a $3\times3\times3$ convolution for local feature extraction, followed by three axis-wise attention modules applied sequentially along three axis. Each block also includes residual connections and layer normalization to stabilize training. Formally, the feature update at block~$t$ is expressed as:
\begin{equation}
F^{(t)} = F^{(t-1)} + \mathcal{A}\!\big(\mathrm{Conv3D}(F^{(t-1)})\big),
\end{equation}
where $\mathcal{A}(\cdot)$ denotes the composite multi-axis attention operation: $\mathcal{A}(F) = 
\mathrm{Attn}_W(
\mathrm{Attn}_H(
\mathrm{Attn}_D(F))).$

After four hierarchical down-sampling stages (stride~$2$ per block), 
the feature map resolution is progressively reduced from the input size $(D,H,W)$ to 
$\left(\tfrac{D}{16}, \tfrac{H}{16}, \tfrac{W}{16}\right),$
with the final channel dimension $C^{4} = 256$. This representation compactly encodes both local anatomical details and long-range structural context, serving as the shared latent feature space for subsequent anatomical and modality-specific branches.

\subsection{Dual-Stream Representation Learning}

To disentangle anatomical structure from modality-dependent appearance, 
our encoder bifurcates into two parallel streams after the shared feature extraction backbone.  
Given the final shared feature map $F^{(4)}$ obtained from the factorized multi-axis attention encoder, 
we define:
\begin{equation}
z_{\mathrm{anat}} = E_{\mathrm{anat}}(F^{(4)}), \qquad
z_{\mathrm{mod}}  = E_{\mathrm{mod}}(F^{(4)}),
\end{equation}
where $E_{\mathrm{anat}}$ and $E_{\mathrm{mod}}$ are lightweight convolutional heads with identical spatial resolution but different learning objectives.

\subsubsection{Vector Quantization of Anatomical Features}
Given the anatomical latent feature map 
$z_{\mathrm{anat}}$, we obtain compact and discrete structural representations $\tilde{z}_{\mathrm{anat}}$ by first introducing the anatomical codebook:
\begin{equation}
\mathcal{C}_{\mathrm{anat}} = \{e_k\}_{k=1}^{K_a}, \quad e_k \in \mathbb{R}^{C_a},
\end{equation}
where $K_a$ denotes the number of codebook entries and $C_a$ is the feature dimensionality. Each spatial location $(i,j,k)$ of  $\tilde{z}_{\mathrm{anat}}$ is then defined by the nearest codebook embedding of the corresponding feature vector $z_{\mathrm{anat}}$, i.e., 
\begin{equation}
\tilde{z}_{\mathrm{anat}}(i,j,k) = e_{k^*} \mbox{ with } 
k^* = \arg\min_{k}\|z_{\mathrm{anat}}(i,j,k) - e_k\|_2^2.
\end{equation}

To ensure that the continuous latent space effectively maps to these discrete entries, we optimize the anatomical stream by minimizing the discrepancy between the encoder's output and its quantized counterpart. Specifically, the vector quantization loss enforces consistency between encoder outputs and their assigned codebook vectors:
\begin{equation}
\mathcal{L}_{\mathrm{VQ}} = 
\|\mathrm{sg}[z_{\mathrm{anat}}] - \tilde{z}_{\mathrm{anat}}\|_2^2
+ \beta \|z_{\mathrm{anat}} - \mathrm{sg}[\tilde{z}_{\mathrm{anat}}]\|_2^2,
\end{equation}
where $\mathrm{sg}[\cdot]$ denotes the stop-gradient operator and $\beta$ controls the commitment strength.  

\subsubsection{Modality-FiLM Decoder}

Given the quantized anatomical codes $\tilde{z}_{\mathrm{anat}}$, the decoder reconstructs the MRI volume through a hierarchy of up-sampling and attention layers. To incorporate modality awareness, each decoder block is modulated using FiLM~\cite{perez2018film} parameters derived from the modality-specific latent $z_{\mathrm{mod}}$. FiLM performs an affine transformation on feature activations, enabling dynamic adaptation of feature intensity and contrast while preserving structural content.

A global modality embedding $s_m \in \mathbb{R}^{C_s}$ (learned per modality) is concatenated with the pooled modality feature:
\begin{equation}
u_m = \mathrm{Concat}\big(\mathrm{GAP}(z_{\mathrm{mod}}),\, s_m\big),
\end{equation}
where $\mathrm{GAP}(\cdot)$ denotes global average pooling.  
A lightweight multilayer perceptron (MLP) then predicts a pair of FiLM parameters $(\gamma_\ell, \beta_\ell)$ for each decoder layer $\ell$:
\begin{equation}
(\gamma_\ell, \beta_\ell) = f_{\mathrm{MLP},\ell}(u_m).
\end{equation}

Given the intermediate decoder feature $h_\ell$, FiLM modulation applies an element-wise affine transformation:
\begin{equation}
h'_\ell = \gamma_\ell \odot h_\ell + \beta_\ell,
\end{equation}
where $\odot$ denotes channel-wise multiplication.  
This operation modulates the feature activations according to modality-specific contrast cues while maintaining anatomical consistency inherited from $\tilde{z}_{\mathrm{anat}}$.

The final 3D reconstruction is expressed as:
\begin{equation}
\hat{x}_m = D\big(\tilde{z}_{\mathrm{anat}}, \{(\gamma_\ell,\beta_\ell)\}\big),
\end{equation}
where $D(\cdot)$ denotes the hierarchical decoder.  
To optimize the decoder for high-fidelity synthesis, we supervise the reconstruction process using a multi-scale objective. Specifically, the reconstruction loss combines voxel-level fidelity and structural similarity:
\begin{equation}
\mathcal{L}_{\mathrm{rec}} = 
\|x_m - \hat{x}_m\|_1 + 
\lambda_{\mathrm{SSIM}}\big(1 - \mathrm{SSIM}_{3D}(x_m, \hat{x}_m)\big),
\end{equation}
where $\lambda_{\mathrm{SSIM}}$ balances the pixel-wise and perceptual terms.

\subsection{Cross-Modal and Disentanglement Losses}

To promote effective disentanglement between anatomical structure and modality appearance, we introduce two auxiliary objectives: a cross-modal reconstruction loss and a modality-adversarial loss.

For paired modalities $(x_{m_1}, x_{m_2})$ of the same subject, we fix the anatomical latent $\tilde{z}_{\mathrm{anat}}^{m_1}$ and swap the FiLM parameters between modalities:
\begin{equation}
\hat{x}_{m_2|m_1} =
D\!\left(\tilde{z}_{\mathrm{anat}}^{m_1}, 
\{(\gamma_\ell^{m_2}, \beta_\ell^{m_2})\}\right).
\end{equation}
The model is then optimized to minimize the reconstruction discrepancy:
\begin{equation}
\mathcal{L}_{\mathrm{cross}} =
\|x_{m_2} - \hat{x}_{m_2|m_1}\|_1,
\end{equation}
encouraging the anatomical code to capture modality-invariant structure while the FiLM parameters encode contrast-specific appearance.

To explicitly remove modality information from the anatomical representation, we introduce a gradient-reversal classifier $D_m$ that predicts the modality label from $z_{\mathrm{anat}}$. The classifier is trained to minimize the prediction loss, while the encoder is trained to maximize it via gradient reversal:
\begin{equation}
\mathcal{L}_{\mathrm{adv}} =
-\mathbb{E}\big[\log D_m(z_{\mathrm{anat}})\big].
\end{equation}
This adversarial objective drives $z_{\mathrm{anat}}$ to be modality-invariant, further enforcing a clean separation between structural and appearance factors.

\subsection{Implementation Details}

Inspired by Joint 2D/3D training in natural video VAEs~\cite{xing2024large} (where alternating between full-frame and image-level inputs improves data efficiency and model generalization), we train our model on 3D brain MRI by mixing the dimensionality of the input. For 2D batches, a random planar slice is sampled from either the axial, coronal, or sagittal view, serving as a data augmentation step. In this case, the rest of axis attentions are disabled, and convolution kernels are adjusted along the selected plane, effectively reducing 3D operations to 2D while preserving in-plane spatial attention. Only the slice-level reconstruction loss $\mathcal{L}^{2D}_{\mathrm{rec}}$ is computed. We employ an exponential moving average (EMA) update for the codebook entries to stabilize optimization and encourage balanced code usage across anatomical regions.

\begin{table*}[t]
  \centering
  \caption{Quantitative comparison with state-of-the-art methods. $f$ denotes the spatial compression ratio of each VAE. All the metrics are higher is better. For each dataset and metric, the best result is highlighted in \textbf{bold}, and the second-best result is \underline{underlined}.}
  \label{tab:compare_sota}
    \resizebox{0.8\textwidth}{!}{\begin{tabular}{cllcccccccc} 
    \toprule
    \multicolumn{1}{l}{}              &                      &     & \multicolumn{4}{c}{T1}           & \multicolumn{4}{c}{T2}            \\ 
    \hline
    \multicolumn{1}{l}{}              & Models               & $f$   & PSNR  & SSIM  & Accuracy & Dice  & PSNR  & SSIM  & Accuracy & Dice   \\ 
    \hline
    \multirow{6}{*}{\rotcelll{NCANDA}} & VQGAN               & 8x  & 26.09 & 90.77 & 76.23    & 90.08 & 25.57 & 88.28 & 66.25    & 89.80  \\
                                      & SD-VAE               & 8x  & 27.48 & 92.72 & 77.14    & 92.51 & 26.16 & 90.87 & 67.34    & 90.37  \\
                                      & MediTok              & 16x  & 28.03 & 94.11 & 77.78    & 93.88 & 27.34 & 93.08 & 67.55    & 92.53  \\
                                      & MedVAE               & 8x  & 28.26 & 94.38 & 78.12    & \underline{94.83} & 27.59 & 92.51 & 67.87    & 92.31  \\ 
    \cline{2-11}
                                      & NeuroQuant (ours)          & \textbf{16x} & \textbf{28.89} & \textbf{95.27} & \textbf{79.09}    & \textbf{95.25} & \textbf{28.12} & \underline{93.57} & \textbf{68.36}    & \textbf{93.45}  \\
                                      & - w/o joint training & \textbf{16x} & \underline{28.51} & \underline{94.89} & \underline{78.57}    & \underline{94.83} & \underline{28.03} & \textbf{93.67} & \underline{67.95}    & \underline{93.03}  \\ 
    \midrule
    \multirow{6}{*}{\rotcelll{ABCD}}   & VQGAN              & 8x  & 24.69 & 89.93 & 77.50    & 89.24 & 24.16 & 87.51 & 66.53    & 88.96  \\
                                      & SD-VAE               & 8x  & 25.94 & 91.92 & 78.37    & 91.65 & 24.69 & 90.04 & 66.58    & 89.53  \\
                                      & MediTok              & 16x  & 26.22 & 93.24 & 79.02    & 93.08 & 25.70 & 91.72 & 66.79    & 91.67  \\
                                      & MedVAE               & 8x  & 26.65 & 93.56 & 79.31    & 93.95 & 26.05 & 92.21 & 67.13    & 91.45  \\ 
    \cline{2-11}
                                      & NeuroQuant (ours)          & \textbf{16x} & \textbf{27.24} & \textbf{94.42} & \textbf{80.34}    & \textbf{94.36} & \textbf{26.53} & \textbf{92.78} & \textbf{67.58}    & \textbf{92.58}  \\
                                      & - w/o joint training & \textbf{16x} & \underline{26.84} & \underline{94.04} & \underline{79.79}    & \underline{94.01} & \underline{26.36} & \underline{92.56} & \underline{67.18}    & \underline{92.16}  \\
    \bottomrule
    \end{tabular}}
    \vspace{-0.5cm}
\end{table*}
\section{Experiments}\label{sec: exp}

In this section, we comprehensively evaluate the proposed NeuroQuant framework. We first describe the experimental datasets, evaluation metrics, and implementation details. Then, we present quantitative and qualitative results, comparing NeuroQuant with several representative VAEs. Finally, extensive visualizations and ablation studies are conducted to analyze the contributions of each component.

\subsection{Experimental Setup}

\noindent\textbf{Datasets} We evaluate NeuroQuant on two publicly available multi-modal 3D brain MRI datasets: The National Consortium on Alcohol and NeuroDevelopment in Adolescence (NCANDA)~\cite{ncanda} and the Adolescent Brain Cognitive Development (ABCD)~\cite{abcd}. Both datasets provide paired T1- and T2-weighted MRI scans, making them suitable benchmarks for assessing multi-modal structural modeling.

\textit{NCANDA} dataset is a longitudinal neurodevelopmental cohort designed to study typical brain maturation and the effects of early alcohol exposure. MRIs were annually acquired for each of the 808 subjects~\cite{pfefferbaum2016adolescent}, providing rich anatomical variability and repeated measurements. We use paired T1- and T2-weighted 3D MRIs from subjects passing the NCANDA QA/QC pipeline, and split the data at the subject level into 606 train, 80 validation, and 122 test subjects, with balanced sex distributions (train/validation/test: 50.5\%/52.5\%/51.64\% female). Reflecting the longitudinal nature of the study, each subject contributes between 1 and 11 visits (\textit{e.g.}, baseline and up to 10-year follow-ups), with an average of 6.25 visits. In total, the training, validation, and test sets consist of 3,790, 489, and 774 MRI sessions, respectively.

\textit{ABCD} study is the largest long-term pediatric neuroimaging cohort to date, aimed at understanding cognitive development and behavioral trajectories. From the released structural MRI data, we extract 8,879 subjects with paired 14,845 T1–T2 3D MRI volumes, following the standard ABCD preprocessing and quality-control pipeline. We split the dataset by subject into train/val/test partitions containing 7,594 / 1,412 / 2,095 unique subjects, corresponding to 11,133 / 1,484 / 2,228 paired MRI volumes. The sex distribution is train/val/test: 52.12\%/53.58\%/51.57\% femalte.

\begin{figure*}[t]
    \centering
    \includegraphics[width=0.9\linewidth]{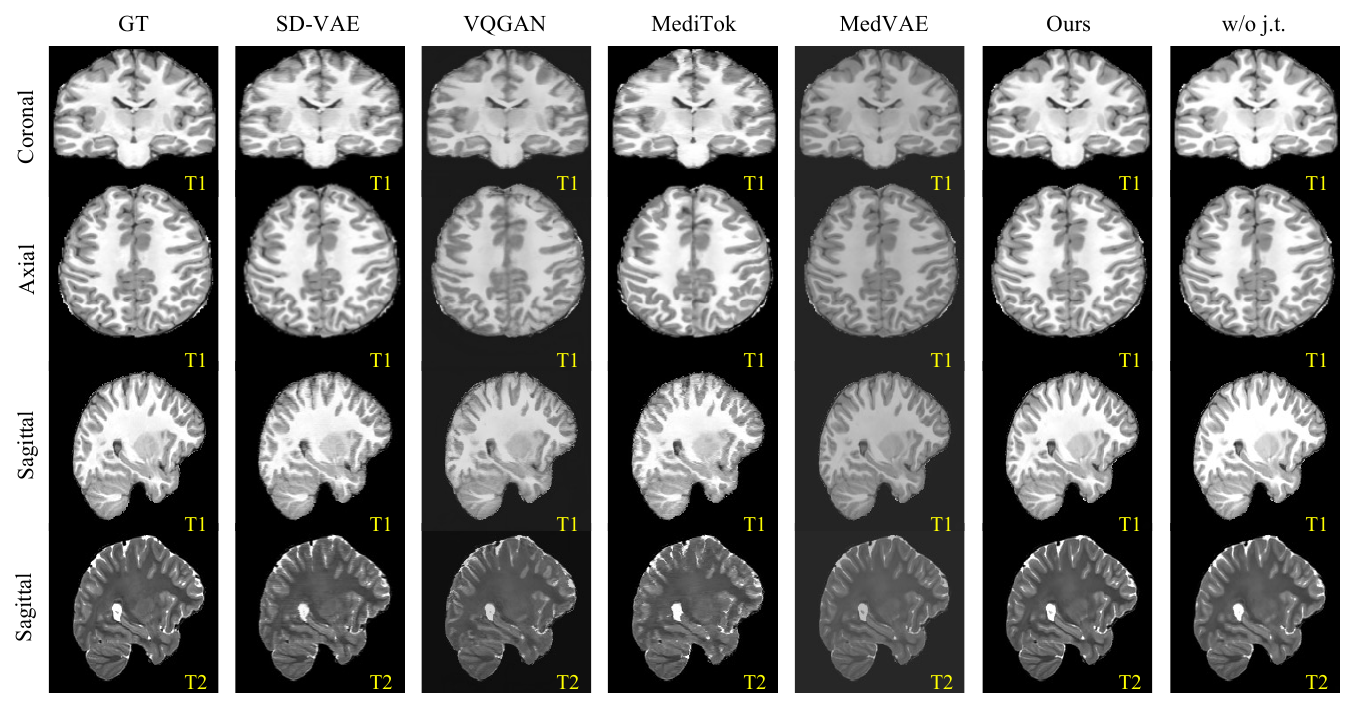}
    \caption{Qualitative comparison of T1- and T2-weighted MRI reconstructions across three views. Unlike existing models that exhibit blurring or structural artifacts (especially in non-axial planes), NeuroQuant achieves superior voxel-level fidelity and anatomical consistency. Notably, our joint 2D/3D training strategy (NeuroQuant) significantly enhances the recovery of fine-grained cortical details and sharp tissue boundaries compared to the 3D-only variant (w/o j.t.).}
    \vspace{-0.5cm}
    \label{fig:qualitiative analysis}
\end{figure*}

\noindent\textbf{Metrics} We evaluate NeuroQuant using three complementary categories of metrics, designed to assess voxel-level fidelity, anatomical consistency, and latent-space quality. Firstly, we use PSNR and 3D SSIM to measure voxel-wise reconstruction accuracy. They provide a direct assessment of how well the VAE restores fine-grained volumetric appearance for each modality. Secondly, following \cite{wang2025toward,wu2024evaluating}, we employ SynthSeg\footnote{\url{https://github.com/BBillot/Synthseg}}~\cite{billot2023synthseg}-based segmentation to evaluate anatomical plausibility. For each reconstructed volume, SynthSeg predicts region-wise brain masks, and we compute the Dice-score compared with the ground truth. This metric reflects the model’s ability to preserve global brain shape, regional volumes, and cross-structure relationships. Lastly, to assess the informativeness of the learned latent space, we evaluate sex classification performance using the MONAI SEResNet-152~\cite{hu2018squeeze} with supervised linear probing trained on frozen latent embeddings. We additionally compare this performance with the same classifier trained directly on the original MRI volumes.

\noindent\textbf{Implementation Details} NeuroQuant is trained on the Marlowe Computing Platform~\cite{kapfer2025marlowe} using Python 3.9 and PyTorch 2.8.0 with a single NVIDIA H100 (80GB) GPU. We merge the training splits of NCANDA and ABCD to form a unified multimodal training cohort. Following 2D/3D joint training strategy in \cite{xing2024large}, we first perform 100k steps of full 3D training using the AdamW optimizer~\cite{loshchilov2017decoupled} with a batch size of 4 and an initial learning rate of $1\times10^{-4}$ that decays to $4.5\times10^{-6}$. We then perform an additional 25k 2D training steps with batch size 128, using a fixed learning rate of $4.5\times10^{-6}$. For preprocessing, each MRI volume is first cropped to the minimal bounding box that fully encloses the brain and then zero-padded to a standardized resolution of [144 (Coronal), 192 (Axial), 144 (Sagittal)] to fit most existing VAEs.

\subsection{Comparison with State-of-the-arts}

\noindent\textbf{Quantitative Analysis} Tab.\ref{tab:compare_sota} reports the quantitative comparison between NeuroQuant and four representative state-of-the-art (SOTA) VAEs, VQGAN~\cite{zhu2024scaling}, SD-VAE~\cite{rombach2022high}, MediTok~\cite{ma2025meditok}, and MedVAE~\cite{varma2025medvae} across both NCANDA and ABCD datasets.
Unlike prior models that operate at an $8×$ spatial compression rate, NeuroQuant achieves $16×$ downsampling, providing a significantly more compact latent representation while still outperforming all baselines. Across both modalities (T1 and T2), NeuroQuant consistently obtains the best or second-best performance on all four evaluation metrics. In voxel-level reconstruction on NCANDA, NeuroQuant surpasses the best baseline (MedVAE) by +0.63 dB PSNR and +0.89 SSIM for T1, and by +0.53 SSIM and +1.14 Dice for T2, demonstrating the benefits of factorized 3D attention and FiLM-modulated decoding in preserving fine-grained anatomical appearance. In structure-level assessment, NeuroQuant achieves the highest SynthSeg Dice scores, confirming its ability to maintain accurate regional brain volumes and volumetric coherence. Performance gains are especially notable on the heterogeneous ABCD dataset, where variations in scanners and acquisition protocols make multimodal disentanglement more challenging. For latent-space evaluation, classifiers trained on NeuroQuant’s latent embeddings achieve the strongest sex classification accuracy on both datasets and modalities.
This indicates that the dual-stream design produces compact yet semantically rich latent codes, superior to KL-based continuous VAEs\cite{rombach2022high,varma2025medvae} and 2D tokenizers such as MediTok, which lack volumetric consistency.

\noindent\textbf{Qualitative Analysis} Fig.\ref{fig:qualitiative analysis} presents qualitative comparisons of reconstructed T1- and T2-weighted MRIs across axial, coronal, and sagittal views. Clear differences emerge among competing VAEs. 2D-based models such as SD-VAE and MediTok exhibit pronounced structural inconsistencies across slices, especially along the coronal plane, where inter-slice dependencies are the weakest. This results in visible artifacts such as discontinuous cortical boundaries, staircase-like distortions, and abrupt contrast changes. These issues highlight a fundamental limitation of 2D VAEs: although effective for natural images, they fail to maintain volumetric coherence required for anatomically plausible 3D brain reconstruction. Such artifacts make them insufficient for downstream neuroimaging tasks that rely on stable 3D structure. MedVAE, while avoiding obvious slice-level artifacts thanks to its 3D convolutional design, tends to oversmooth anatomical details. Fine structures such as cortical folding, sharp ventricular boundaries, and thin white-matter interfaces become noticeably blurred. This smoothness reflects a limitation of KL-based VAEs, which often favor reconstruction stability at the cost of high-frequency anatomical textures. In contrast, NeuroQuant produces clearly sharper edges, smoother inter-slice transitions, and anatomically coherent structures across all three orientations. Cortical folds are preserved with higher fidelity, deep gray nuclei maintain consistent shapes, and ventricular boundaries remain sharp without collapsing or over-smoothing.

\subsection{Ablation Study}

\noindent\textbf{Effect of 2D/3D Joint Training(j.t.)} We evaluate the impact of the adapted 2D/3D joint training strategy by comparing NeuroQuant with and without this component. Joint training alternates between full 3D volumes and randomly sampled 2D slices, allowing the model to learn both volumetric structure and high-frequency slice-level details. Quantitatively, removing joint training consistently degrades performance across all metrics (Tab.~\ref{tab:compare_sota}).
On NCANDA, the full model improves T1 reconstruction by +0.38 dB PSNR, +0.38 SSIM, and +0.42 Dice, and increases latent sex classification accuracy by +0.52\%.
On the more heterogeneous ABCD dataset, gains are even more pronounced. These improvements demonstrate that joint training enhances both low-level fidelity and high-level anatomical coherence. Qualitatively (Fig.\ref{fig:qualitiative analysis}), the model without joint training exhibits blurrier edges and mild slice inconsistencies, particularly in coronal views. This suggests reduced generalization across orientations and weaker anatomical texture recovery. Overall, these results show that 2D/3D joint training is essential for achieving sharper reconstructions, stronger volumetric coherence, and more informative latent representations.

\begin{table}[t]
\centering
\caption{Ablation study on the NCANDA dataset evaluating the contribution of key components in NeuroQuant. We incrementally add factorized multi-axis attention (FMA), modality-FiLM decoder (M-FiLM), $\mathcal{L}{\mathrm{cross}}$, and $\mathcal{L}{\mathrm{adv}}$.}
\label{tab:ablation study}
\resizebox{0.4\textwidth}{!}{\begin{tblr}{
  column{2} = {c},
  column{3} = {c},
  column{5} = {c},
  column{6} = {c},
  cell{1}{2} = {c=3}{},
  cell{1}{5} = {c=3}{},
  cell{3}{4} = {c},
  cell{3}{7} = {c},
  cell{4}{4} = {c},
  cell{4}{7} = {c},
  cell{5}{4} = {c},
  cell{5}{7} = {c},
  cell{6}{4} = {c},
  cell{6}{7} = {c},
  cell{7}{4} = {c},
  cell{7}{7} = {c},
  hline{1,8} = {-}{0.08em},
  hline{2-3} = {-}{},
}
              & T1    &       &          & T2    &       &          \\
Settings      & PSNR  & SSIM  & Accuracy & PSNR  & SSIM  & Accuracy \\
Base model    & 27.56 & 92.18 & 77.38    & 26.43 & 91.24 & 67.51    \\
+ FMA         & 28.17 & 94.37 & 78.22    & 27.39 & 92.08 & 67.85    \\
+ M-FiLM     & 28.52 & 94.84 & 78.39    & 27.83 & 92.49 & 68.01    \\
+ $\mathcal{L}_{\mathrm{cross}}$     & 28.63 & 95.01 & 78.45    & 27.94 & 92.76 & 68.06    \\
+ $\mathcal{L}_{\mathrm{adv}}$ (full) & 28.89 & 95.27 & 79.09    & 28.12 & 93.57 & 68.36    
\end{tblr}}
\vspace{-0.5cm}
\end{table}

\noindent\textbf{Effect of Key Modules(k.m.)} Along with the joint training strategy, we further perform ablation studies beginning with a pure 3D VQ-VAE as our base model and progressively introduce each component to assess its individual contribution. The performances are presented in Tab.\ref{tab:ablation study}. Introducing FMA leads to a substantial improvement over the base model, boosting T1 PSNR from 27.56 to 28.17 and SSIM from 92.18 to 94.37, and yielding similar gains on T2. These results demonstrate that FMA effectively enhances long-range structural modeling and improves the reconstruction of global brain geometry. Adding the M-FiLM decoder further increases reconstruction quality and latent discriminability.
M-FiLM improves T1 SSIM to 94.84 and raises sex classification accuracy from 78.22\% to 78.39\%, indicating that modality-adaptive modulation sharpens fine-grained anatomical textures while preserving structural coherence. Together, FMA and M-FiLM enhance both voxel-level fidelity and modality-aware detail reconstruction.

\begin{figure}
    \centering
    \includegraphics[width=1\linewidth]{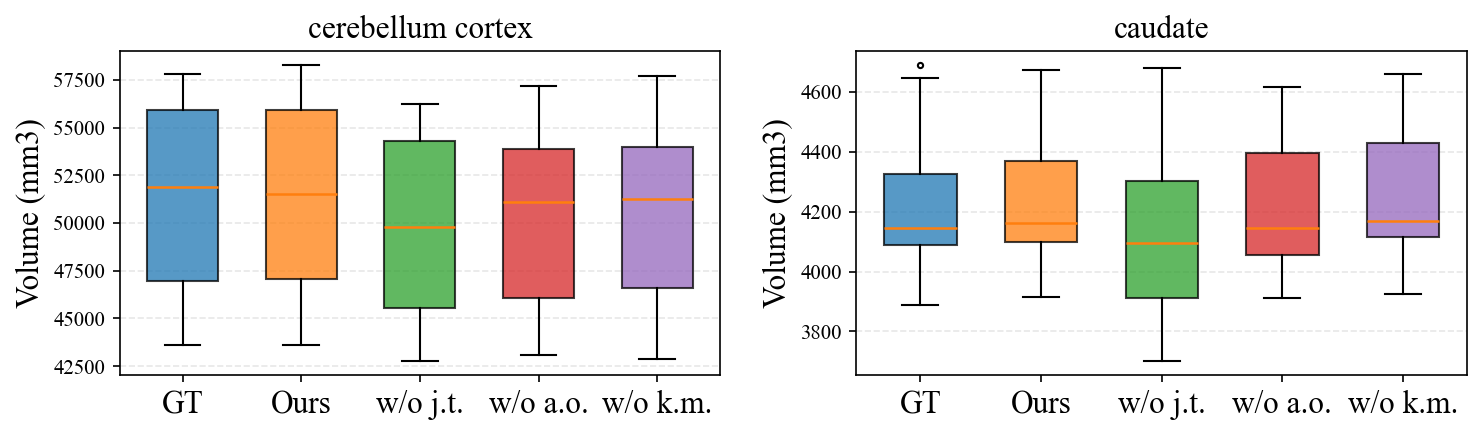}
    \caption{SynthSeg-based cerebellum cortex and caudate volume comparison between GT, our full model, and ablated variants.}
    \vspace{-0.5cm}
    \label{fig:region_differences}
\end{figure}

\noindent\textbf{Effect of Auxiliary Objectives(a.o.)} Beyond architectural modules, we evaluate the role of the two disentanglement objectives: the cross-modal reconstruction loss $\mathcal{L}_{\mathrm{cross}}$ and the modality-adversarial loss $\mathcal{L}_{\mathrm{adv}}$. Adding $\mathcal{L}_{\mathrm{cross}}$ consistently improves PSNR, SSIM, and accuracy across modalities, confirming that explicitly aligning paired T1–T2 reconstructions encourages the anatomical code to remain modality-invariant. The addition of $\mathcal{L}_{\mathrm{adv}}$ yields the full NeuroQuant model, which achieves the best performance across all metrics (e.g., T1 PSNR 28.89, SSIM 95.27, Accuracy 79.09). This demonstrates that adversarial removal of modality cues from the anatomical latent representation further strengthens the disentanglement of structure and appearance, leading to cleaner anatomical codes and more discriminative latent features.

To further assess anatomical consistency, we compare region-wise brain volumes estimated by SynthSeg across different ablation settings (Fig.\ref{fig:region_differences}). For both the cerebellum cortex and caudate, our model produces volume distributions that closely match the ground-truth volumes. Removing key components, such as joint training, auxiliary objectives, or key modules, leads to larger deviations and increased variability. These results indicate that each module contributes to maintaining anatomically faithful structure, and the full model achieves the highest level of structural reliability.

\section{Conclusion} 
We presented NeuroQuant, a unified 3D VQ-VAE designed to jointly model anatomical structure and modality-specific appearance in multi-modal brain MRI.
Through a dual-stream encoder, factorized multi-axis attention, and a modality-aware FiLM decoder, our framework learns disentangled and anatomically faithful latent representations while achieving a compact $16\times$ spatial compression. A 2D/3D joint training strategy further enhances reconstruction quality and volumetric coherence. Extensive experiments on NCANDA and ABCD demonstrate that NeuroQuant consistently outperforms existing VAEs in voxel-level fidelity, structure-level consistency, and latent-space discriminability. Our results highlight the value of unified multimodal encoding for scalable brain MRI modeling and provide a solid foundation for future generative and analytical neuroimaging applications.

\section*{Acknowledgment} 
This work was partly supported by the National Institute of Health (AA021697, DA057567), and by the Stanford University Human-Centered Artificial Intelligence (Google Cloud Credit, Hoffman-Yee Award).
\thanks{NCANDA data collection and distribution were supported by NIH funding AA021681, AA021690, AA021691, AA021692, AA021695, AA021696, AA021697, AG089169. They are made publicly accessible via \url{https://nda.nih.gov/edit_collection.html?id=4513}}

{
    \small
    \bibliographystyle{ieeenat_fullname}
    \bibliography{main}
}


\end{document}